# Artificial Intelligence Ecosystem for Automating Self-Directed Teaching

Tejas Satish Gotavade
*Computer Engineer*
Terna Engineering College, Navi Mumbai
Panvel, India
gotavadetejas2122@ternaengg.ac.in

*Abstract*—This research introduces an innovative artificial intelligence-driven educational concept designed to optimize self-directed learning through personalized course delivery and automated teaching assistance. The system leverages fine-tuned AI models to create an adaptive learning environment that encompasses customized roadmaps, automated presentation generation, and three-dimensional modelling for complex concept visualization. By integrating real-time virtual assistance for doubt resolution, the platform addresses the immediate educational needs of learners while promoting autonomous learning practices. This study explores the psychological advantages of self-directed learning and demonstrates how AI automation can enhance educational outcomes through personalized content delivery and interactive support mechanisms. The research contributes to the growing field of educational technology by presenting a comprehensive framework that combines automated content generation, visual learning aids, and intelligent tutoring to create an efficient, scalable solution for modern educational needs. Preliminary findings suggest that this approach not only accommodates diverse learning styles but also strengthens student engagement and knowledge retention through its emphasis on self-paced, independent learning methodologies.

## I. Introduction

The Artificial Intelligence is currently being researched by on a great scale and continuously growing drastically these last few years. Many new businesses are adapting and utilizing the power of Artificial Intelligence in order to scale their technologies. Nowadays, almost every application is designed to utilize the concept and provide faster and accurate results.

Education is a crucial sector which is growing day by day with astonishing developments seen in new research and overall study material collection. It has become easier to discover new concepts and learn them just by connecting to internet. Anyone can now gain knowledge in their own pace and focus on working what they actually desired to work instead of being confused on what to do and what not. Right now, education industry has become so vast that it creates a hurdle to provide a student with efficient knowledge at low cost, student focused curriculum, restricted pace of understanding with deadlines. We start studying from our childhood on a specific curriculum designed on generalised importance of knowledge and not what the student wants to learn.

### A. Importance of Self Learning

The landscape of educational technology continues to evolve rapidly in response to changing professional demands and learning paradigms. As [4] Curran et al. (2019) emphasize, "A fundamental principle of continuing professional education is lifelong learning, whereby professionals pursue knowledge and skills to maintain competency in their field of practice." This foundational concept underscores the critical need for flexible, technology-enhanced learning solutions that support continuous professional development. In today's digital age, traditional educational systems face unprecedented challenges in meeting these evolving needs for lifelong learning and professional competency maintenance.

Based on professional learning, we need to be able to open to changes happening in our surrounding and use it to figure out optimal destination. Guided paths end up making a person more vulnerable to changing environments as a small discomfort could lead to greater cause.

### B. Knowledge Perspective

Internet is like a void with ever-increasing data and complex architecture. Throughout the world, it is being used all the time for day to day tasks like chatting, sharing pictures online, taking virtually and the list goes on. Searching through the internet results answers for the questions asked but what should we ask? If I want to discover a concept how much should be learned? Will the answers be enough or is there more to it? Figuring out the right path is a tedious task and can end up making someone frustrated and giving up for the time being on what they wanted to do. This problem arises due to large data and less context specificity, which ends up with endless connections with nodes of knowledge attached to each other going on and on.

To overcome this barrier, we can adapt to latest technologies under Artificial Intelligence. All the data can be fitted into a Machine Learning model in order to process it according to required needs of student or knowledge enthusiast.

In this paper, we will be working on developing a new concept and implementing it. Our focus will be creating a software that aims on what the user will actually require and assisting it using the *Self-Directed Teaching* architecture which will help to fill the gap between human interaction and understanding between machines. This will assist the user to achieve undiscovered potential while understanding new knowledge at own desired pace.

Our concept introduces a new logical flow of conversation which will be termed as *Self-Directed Teaching*. We will be creating a directed path of architecture that will align the user to gain context specific knowledge efficiently. The data will be collected throughout internet dynamically and cleaned before utilizing order to provide accurate and precise educational content.

### C. Combining Concepts

Now we got a glimpse of what is the current condition of knowledge and how could it be pursued efficiently, we move forward to combing and forming a new concept that will provide the power of Artificial Intelligence while preserving the importance of Self learning.

The two terminologies will be connected in a loop in order to provide an optimal output of desired thinking. This will generate a technology which could be used to gain context-specific knowledge from filtering out the massive collection of data in Internet and process it according to depth of requirements given to it. The user will now be able to interact with machine in a virtual environment with more awareness of situation assigned to it. This type of advancement will develop a distraction free surrounding in order to let the user pursue its path based on variable speed of understanding and learning.

This concept is designed in order to tackle the ever-growing supply of scattered knowledge data over Internet and process it in such a way that user gets the results which should have been achieved.

## II. OBJECTIVE

We emphasise on developing a software that will join the broken bridge between student and a professor. Mostly in schools or colleges, we often end up getting less attention due to number of students present in surrounding. There seems be no way of getting assistance when required. Many online platforms have started building business scheme over this idea with providing one to one real-time support at the price of college fees for a year. As a student, it is becoming difficult to focus on goals as we end up sticking to predefined curriculums which we all are working on. Many students get stuck in the loop and end up never finding their true goal. This fact could also potentially be the cause of unemployment as students are trained on same things rather then what they want, leading to saturation in a single sector. If there could be a way to understand and peruse knowledge based on self-awareness and actual interests then a lot of problems existing in surrounding could have been solved.

The goal of this article is to showcase the concept of Automated Self learning that will create an environment where students can learn new concepts in a more controlled manner and discover accurate as well as adequate path that will lead to optimal learning. We will be developing an architecture that will promote *Self learning and Automation in Teaching.* This idea will be designed to target mainly on students studying in schools or colleges who are looking for ways to increase their knowledge and *knowledge enthusiasts* or people that want to peruse in depth knowledge in a particular field of study.

Utilizing knowledge at an affordable and restriction free environment will help the students in understanding specific fields in depth. This will also improve an individual's decision making as they can start taking their own choices for opting for courses of their likings and pursuing them. This concept is aligned to work on various fields such as psychological improvement, focussing environment, real-time doubts solving and many more. We will be talking about this in detail in further sections.

## III. GAP IDENTIFICATION

There are a lot of problems present in the current education industry. Many students face these problems in their day to day life but end up ignoring them. These problems listed are the ways students fail securing their desired goal.

### 1. No personal assistance

In a daily tight schedule of schools and colleges, the main focus is to finish the syllabus. Everyone is focused on securing grater grades in order to secure a good college or get placed in a job. This lifestyle ends up misleading many valuable students from achieving their actual potential. There is no guidance provided as the giving personal assistance is a hard work and too much time consuming. This leads to students pursuing wrong degree and working in uninterested jobs.

### 2. Restricted Environment

Students are trained to learn topics under restricted conditions. Every student has a different pace of understanding new concepts and adapting to work. Since, the curriculums are designed same for everyone, many undirected students end up failing. This is the result of fixed deadlines assigned to a student.

### 3. Context Specific Knowledge

There are two types of websites that fulfil the criteria of educating an individual: Informative websites and Pre-recorded teaching. In informative websites, we get the required amount of accurate data to learn but there is no one to explain or solve doubts. This leads to losing interest in something you want to learn due to understanding barrier. In Pre-defined recorded videos, we get sufficient data just to understand the concept but again solving real-time doubts are an issue. There is also the teaching tutors that teach a student on specific context but the drawback is that the fees required for a single subject is not affordable to all.

### 4. Distractions

Consider a case of a student studying in a class filled with 50 students. In the class, we end up having useless conversations, group discussions, peer pressured decisions, and restricted choice selection. If a student is in a lower grade then we could ignore these factors but in case of a grown student, these factors are crucially scalping its future. This environment restricts the student from thinking beyond the box and check if there is a possibility or not.

### 5. Un-guided path of teaching

In a group of students pursuing same path have same set of curriculums assigned. In case, a student wants to learn something different, they end up getting lost in poorly guided YouTube tutorials or generalised paths leading to loss of interest.

### 6. Expensive Courses

When a student decided to learn some specific knowledge, mostly its first choice is to search for a course online. There are many websites that are build to cover this need. The problem comes when the student is unable to purchase them as a context specific course is costly. Some companies also charge monthly subscriptions similar to streaming platforms which also makes the student burdened with course expiration.

### 7. Automated Generated Content

There is still lack of automated content generation software designed due to complex architecture and domain specific convergence. Most of the business under educational sector are generalized leading to *excess information generation* problem. Using Large Language Models (LLMs) such as ChatGPT or ClaudeAI are developed on generalized dataset and generates extra or un-necessary details which would not be required for a student.

These are just the most important problems listed as the list of problems goes on. We will be focusing on tackling the current problems in the education sector by creating an architecture that will solve these issues. With the use combining the concept of Self Learning and Artificial Intelligence, we will provide a user interface where user can freely opt for courses and get guided path towards accurate and precise learning path without restricted timeline attached.

## IV. PSYCHOLOGICAL FOUNDATION

The foundation of this project is mainly based on the psychological concept of better understanding from self-learning. This part is highlighted as it is a crucial factor that needs to be separately dealt with. *Self-learning* is the terminology that states that pursuit of knowledge should not be driven by culture or external pressure. This idea is implemented for *promoting self-awareness and managing change to surrounding*.

### A. Benefits of Self Learning

There are many advantages of self-learning as it is the concept that makes an individual more conscious being. Some of the pros on this concept are:

- *Flexibility*: Self-learning gives learners the freedom to learn at their own pace and in a comfortable environment.
- *Confidence*: Self-learning can help learners build confidence by overcoming obstacles and learning from their mistakes.
- *Independence*: Self-learning encourages learners to be independent and actively engage with the material.
- *Problem-solving*: Self-learning can help learners develop problem-solving skills.
- *Time management*: Self-learning can help learners improve their time management skills.
- *Broadened thinking*: Self-learning can help learners broaden their thinking level.
- *Self-esteem*: Self-learning can boost a learner's self-esteem by allowing them to go beyond what their teachers or textbooks offer.
- *Work ethic*: Self-learning can help learners develop a good work ethic.
- *Retention*: Self-learning can help learners retain information more naturally.
- *Accessibility*: Self-learning can be accessible for people in remote areas.

### B. Cognitive advantages of personalized learning paths

The *cognitive advantages* of personalized learning paths refer to the mental benefits gained when educational experiences are tailored to meet individual learning needs, preferences, and abilities. This approach enhances engagement, improves memory retention, reduces cognitive load, and fosters self-directed learning, leading to a deeper understanding of content and more effective skill acquisition. Through personalization, learners can progress at their own pace and focus on areas where they need improvement, supporting overall cognitive growth and lifelong learning abilities.

Personalized learning paths offer several cognitive advantages by adapting to the needs, abilities, and interests of each learner. Here are some key benefits:

- *Enhanced Engagement and Motivation*: Personalized learning aligns content with the learner's interests, promoting intrinsic motivation. When students find material relevant, they're more likely to be attentive, resulting in improved retention and comprehension.
- *Improved Memory Retention*: Customizing learning materials to suit individual preferences can enhance memory retention. For instance, visual learners might retain information better with graphics, while others may prefer textual or auditory content.
- **Efficient Problem-Solving Skills**: Personalization allows students to learn at their own pace and explore problem-solving methods that work best for them. This can lead to a deeper understanding of concepts, fostering critical thinking and creativity.
- **Reduction of Cognitive Load**: By focusing on areas where learners struggle and skipping concepts they already understand; personalized paths reduce unnecessary cognitive load. This prevents mental fatigue, making it easier to absorb new information.
- **Self-Directed Learning and Metacognition**: Personalized learning encourages learners to take responsibility for their progress, enhancing metacognitive skills. As they assess their strengths and weaknesses, they develop strategies for effective learning, which can benefit lifelong learning.
- **Better Skill Development and Application**: Tailoring education to individual learning styles enables learners to acquire skills that they can apply practically, as they see immediate relevance to their lives and goals.

From the above, we can say that personalized learning paths support cognitive development by fostering motivation, enhancing retention, and enabling self-directed learning, resulting in more effective and enjoyable learning experiences.

## V. LITERATURE SURVEY

Let's start our research from initial logic and advance further preserving the base of knowledge learned from it.

### A. Historical background and Methodologies on Self-learning

As [1] Barry J. Zimmerman's article discusses the evolution of research on self-regulated learning (SRL) and its relationship with motivation in academic settings. The paper highlights methodological innovations, findings from trace logs of SRL processes, and the implications for educational practices. It is defined as the self-directive processes and self-beliefs that enable learners to transform their mental abilities into academic performance skills. It emphasizes proactive strategies such as goal setting, strategy selection, and self-monitoring. The article discusses the development of innovative tools like the gStudy software, which allows learners to engage in various self-regulatory activities, including note-taking, concept mapping, and collaborative learning. This software captures detailed trace logs of students' SRL processes, providing valuable data for researchers.

A significant question raised is the comparison between trace measures of SRL and self-report measures. Research by Winne et al. (2006) indicates that trace measures can provide a more nuanced understanding of students' self-regulation during learning cycles.

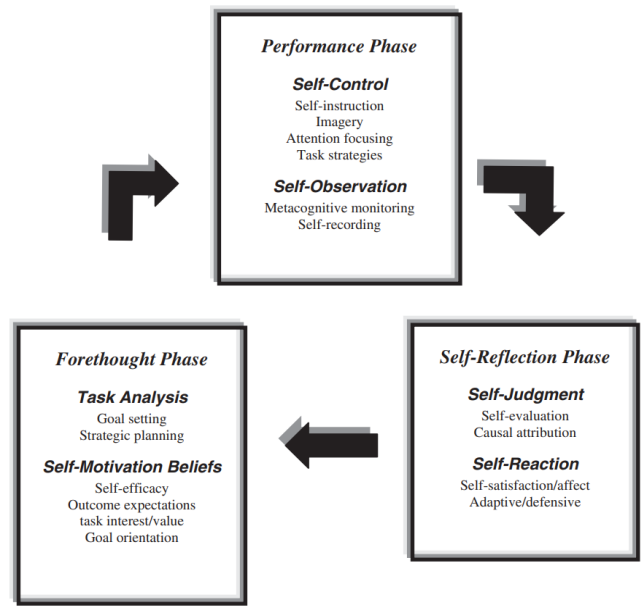

*Fig. Phases and sub processes of self-regulation. From "Motivating Self-Regulated Problem Solvers" by B. J. Zimmerman and M. Campillo, 2003, in J. E. Davidson and R. J. Sternberg (Eds.), The Nature of Problem Solving, p. 239. New York: Cambridge University Press. Copyright 2003 by Cambridge University Press. Adapted with permission. [1]*

[1] Zimmerman calls for further research to refine online measures of SRL for various academic contexts and to explore the effectiveness of interventions aimed at enhancing students' motivation and self-regulation. Finally, the article concludes that while traditional aptitude measures of SRL are valuable, online event measures offer deeper insights into the interrelations of SRL processes in real-time. Understanding these dynamics is crucial for diagnosing and addressing self-regulatory challenges in learners.

### B. Self-Awareness Development

**Self-awareness** is the ability to see oneself clearly, encompassing both internal (affect, beliefs, values) and external (perceptions of others) components. It is crucial for personal and professional growth, influencing interpersonal relationships and work behaviours.

Here, in the [2] paper, we find that the review focused on studies from 2016 to 2021, identifying 42 articles that measured self-awareness changes in college students and working adults. The literature highlights a variety of educational interventions aimed at enhancing self-awareness, including: *Mindfulness Training*: Encourages present-moment awareness and acceptance. *Reflective Practices*: Promotes critical thinking about personal experiences. *Coaching and Feedback*: Provides insights into personal strengths and areas for improvement. *Leadership Development Programs*: Enhances self-awareness among leaders, leading to better performance.

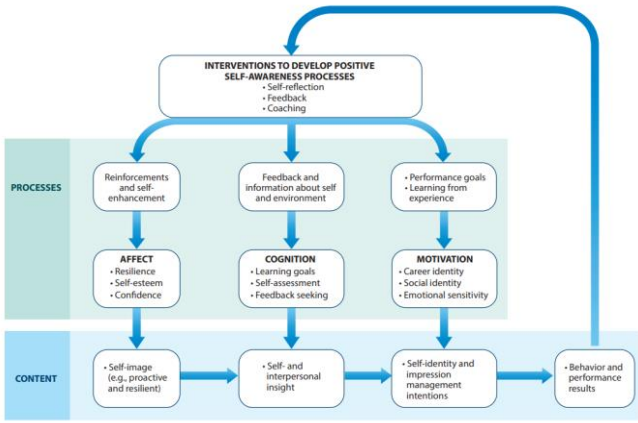

*Fig. An integrative model of positive and continuous self-awareness development through self-reflection, feedback, and coaching. [2]*

The paper also synthesizes affective, cognitive, and motivational processes, emphasizing the role of situational support in developing resilience and self-identity. The review suggests that societal norms and organizational culture significantly influence self-awareness development. Self-presentation on social media reflects self-awareness and can impact personal branding and identity.

Now we can conclude that, it emphasizes the multifaceted nature of self-awareness development and the need for continued research to refine interventions and measurement tools. Future studies should focus on integrating theoretical frameworks with practical applications to enhance self-awareness in various settings.

### C. Educational Advantages from Self-Information processing

The literature on self-related information processing highlights its significance in educational contexts, particularly in enhancing student motivation, engagement, and performance. This review synthesizes key findings from various studies that explore the interplay between self-relatedness and learning outcomes.

Utility value interventions encourage students to connect academic content to their personal lives. Studies [2] (Hulleman & Barron, 2016; Harackiewicz et al., 2016) demonstrate that such interventions significantly improve engagement and interest, especially among students with low performance expectations. By making personal connections, students find relevance in the material, which enhances their intrinsic motivation and perceived value of the content.

Research indicates that utility value interventions are particularly beneficial for underrepresented minority (URM) and first-generation (FG) students (Harackiewicz et al., 2016). Studies [2] (., Schilbach et al., 2012) suggest that self-related processing activates reward pathways in the brain, which may explain the increased motivation and performance observed in students who engage in self-referential tasks. The intersection of educational psychology and neuroscience provides a comprehensive understanding of how self-relatedness influences learning.

Personal connections made by parents regarding school subjects can enhance children's motivation and interest in learning [2] (Walkington et al., 2013).

The paper underscores the importance of self-related information processing in education. By implementing utility value interventions and fostering personal connections, educators can enhance student engagement and performance, particularly among diverse populations. Future research should continue to explore the mechanisms behind these effects and the potential for integrating neuroscientific findings into educational practices.

### D. Can Large Language Models (LLMs) be used for teaching

Traditional teaching literature emphasizes small-scale tasks and analytical models. The primary aim is to enhance reasoning abilities in state-of-the-art large language models (LLMs).

Methods which could be used here are as follows: *Regularization Loss*: An independent study introduces a regularization loss to improve model generalization, filtering out extraneous details for better imitation by auxiliary students. *Math-Shepherd*: This method proposes an automatic scoring system for partial rationales, evaluating how effectively a "completer" model can derive correct answers from these rationales.

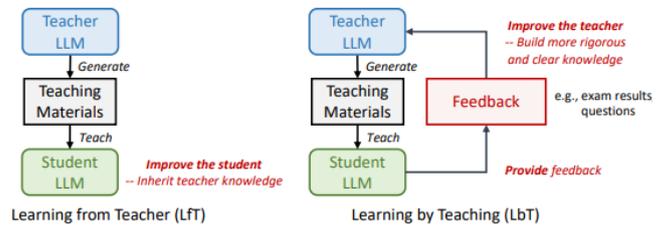

*Fig. Learning by Teaching (LbT) [5]*

The Learning by Teaching (LbT) demonstrated in [5] states the approach that shares conceptual ground with the assumption that generalizable correlations are easier to imitate. Two prompting-based methods are explored, focusing on enhancing the output generation pipeline and optimizing prompts iteratively.

From this paper we can say that the research paper effectively addresses the enhancement of reasoning abilities in large language models through innovative methodologies, including Learning by Teaching (LbT) and regularization techniques. It provides a comprehensive literature review, articulating the theoretical foundations and empirical results while acknowledging limitations and ethical considerations. Overall, the findings contribute valuable insights to the field, paving the way for future advancements in model performance and societal impact.

### E. Fine Tuning Large Language Models (LLMs) for Self-Teaching

The research focuses on the development of **Tailor-Mind**, an intelligent visualization system designed to enhance self-regulated learning (SRL) through the integration of fine-tuned large language models (LLMs) and visualization techniques. The study emphasizes the importance of effective learning tools that support students in navigating complex knowledge areas and improving their learning outcomes.

The study highlights the significance of SRL in education, advocating for models like [6] Zimmerman's SRL framework to foster student initiative and engagement. Participants reported that the self-reflection phase of learning helped them identify misunderstandings and set clear learning goals.

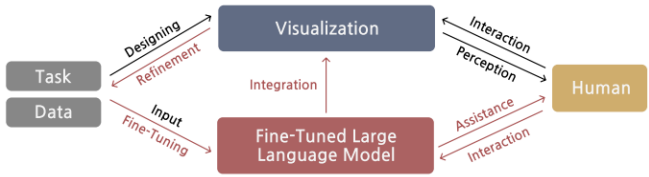

*Fig. Framework of integrating fine-tuned LLM into visualization system. [6]*

The system provides personalized question recommendations and learning paths, aiding users in exploring knowledge connections. It incorporates visual aids, such as knowledge structures and mind maps, to reduce cognitive load and enhance comprehension. The system encourages active exploration and inquiry, making learning more interactive and enjoyable.

User studies indicated that Tailor-Mind significantly improved performance on objective experimental questions, with positive feedback on usability and effectiveness. Participants appreciated the clarity of explanations and the inclusion of examples, which facilitated understanding of complex concepts. The system's design minimizes cognitive load by providing precise and comprehensive responses, particularly beneficial for novice learners. Participants noted that the structured approach to learning helped them integrate fragmented knowledge into a cohesive understanding.

The [6] paper presents Tailor-Mind as a promising tool for enhancing self-regulated learning through intelligent visualization and personalized recommendations. The findings underscore the importance of integrating technology in education to support learners in navigating complex knowledge landscapes and achieving their learning goals.

### F. Fine Tuning Large Language Models (LLMs) for Enterprise

The [7] paper explores the fine-tuning of Large Language Models (LLMs), specifically LLaMA, using proprietary documents and code repositories. It emphasizes the importance of preparing training data effectively to enhance model performance and reduce hallucinations during inference.

LLMs can learn from limited data with appropriate configurations, but hallucinations remain a challenge. Experimentation with different prompt templates and chunking techniques, such as semantic chunking, is suggested to mitigate hallucinations. Fine-tuned models demonstrated improved response quality, particularly in following document styles and answering complex queries.

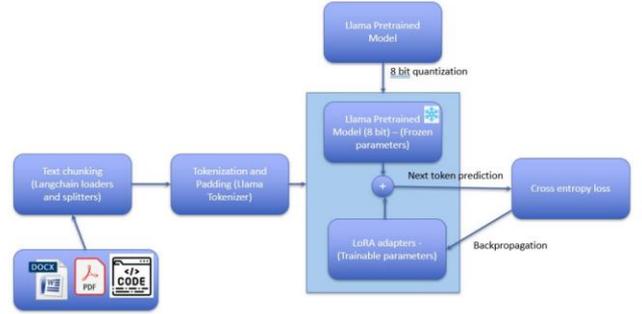

*Fig. Fine tuning Workflow [7]*

The [7] paper provides practical recommendations for beginners on fine-tuning LLMs, including: Estimating GPU requirements and memory needs. Choosing optimal dataset formats and configurations for fine-tuning. Utilizing efficient techniques like Low Rank Adaption (LoRA) to minimize hardware requirements while maintaining performance.

From this [7] paper we learn that the research highlights the significance of tailored dataset preparation and fine-tuning strategies for LLMs in enterprise applications. It underscores the demand for domain-specific models due to privacy concerns and the need for high-quality responses in various use cases, such as support ticket resolution and document querying.

## VI. METHODOLOGY

Now we can get into the technical aspect of the concept proposed through this article. We will be using multiple domain integrations in order to reach the expected goals approximated from psychological as well as technical aspects.

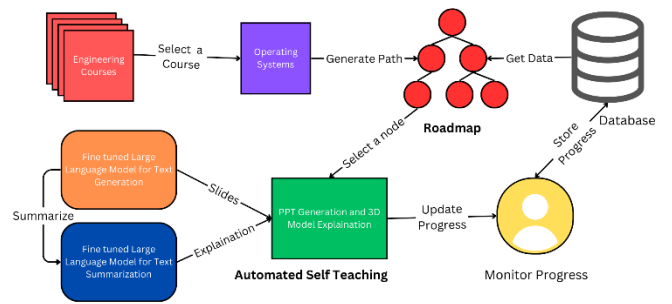

*Fig. Concept Architecture*

### A. Core Components

There are multiple stages of work-flow required in order to create such architecture.

- *AI-Powered Course Roadmap Generation:* The implementation of AI-driven course roadmap generation represents a significant advancement in personalized learning methodologies. This system employs machine learning algorithms to analyse student performance data, learning patterns, and educational objectives to create customized learning pathways.

- *Auto-generated Presentation:* The system operates through a sequential combination of multi-stage processing under the logic. The user will be provided to select a node/topic from the roadmap displayed for a particular course. The node will lead to our custom fine-tuned Large Language Model (LLM) processing which will take an input as the topic and result it in page by page creation of static presentation template covering all the possible areas of understanding feasible under the topic. The text output will be combined in order to demonstrate a Power Point Presentation (PPT) like appearance for visual demonstration.

- *Summarization Model:* After the text generation from the fine-tuned LLMs, the text will be sequentially updated to our fine-tuned Summarization LLMs. This model will help to summarize and create shorter as well as accurate words of quality. The purpose of summarization is to take the generated output of shorter text in order to provide it to the 3D-model for speaking and explanation.

- *3D-Model:* We aim to design a virtual professor which will act as an assistant for the user. It will be used for explanation and doubt solving entity. The summarized text will be inserted as input into the 3D-Model and using a third-party Application Programming Interface (API), it will speak naturally as similar as a human speaks. The model will also capable of solving doubts through 2 possible ways;

  - *Chat System:* User can chat with the context specified fine-tuned LLMs and extract specific info on a topic through text.

  - *Voice Assistance:* User has also the ability to ask the professor by calling an allotted key-word that will trigger the input of voice and process the output through the 3D-Model.

- *Voice Assistant:* In order to create an actual virtual studying environment, a voice assistance will be provided in order to speak with the professor more dynamically. Using the allotted professor name or some key-word, we could call the auto-generation to take a pause and enable the virtual professor to solve any queries. The aim of this component is to provide a one-to-one interaction with Artificial intelligence through the mode of speech in order to provide reality and engagement to the concept.

All these components play an important role towards designing a software that will help a student utilise professional learning with higher focus and less distraction.

### B. Fine Tuning Large Language Models (LLMs)

Let's start with the main component development and actual ways of achieving optimal results. We will be using multiple LLMs in order to test which will be best for different scenarios. The flow of process starts as follows:

1. *Data Collection*: In order to fine-tune a Large Language Model, we need to collect a large set of context specific data. The data is an important part of this system as accurate and precise data is our goal in the long term. The steps involved in data collection are:
   a. *Data Scraping*: We will be using data scraping in order to extract relevant text from the internet. Using python's beautiful soup library, we will scrap the data from various books, educational websites and context specific educational providers.
   b. *Data Cleaning*: It helps in removing the errors and empty values in a dataset. This make the data ready to use for processing and transforming.
   c. *Data Merging*: After filtering out precise and positive data, we will be combining the all the similar sources of data together in order to generate a larger dataset. This is done in order to provide a larger version of data to Large Language Model in order to tackle the under fitting issue.
   d. *Data Organization*: Now we use the collected data and split it into subsets of training, testing and validation. These will be used as input while fine-tuning a Large Language Model (LLM).

2. *Model Selection:* Now that we have the dataset, we can proceed to develop the model. Each model will be developed separately in order to keep uniqueness and context of the topic.

**Fine Tuning Process**

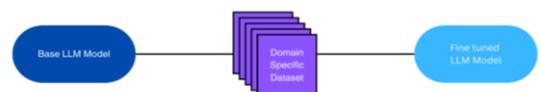

*Fig. Fine Tuning Procedure*

The Base Model here are, *Llama 3.1 8B, Mistral-Nemo-Base-2407 12B, Phi 3.5 mini-instruct and Gemma 9B* where B stands for billion parameters. Each model will be trained on custom dataset to compare accuracy among different domains. We will be using the 4-bit quantization as large models require more hardware and GPU to process. It helps to preserve almost all information by scaling it to lower resources.

3. ***Optimization of Large Language Model (LLM):*** In order to get the best out of the model while keeping the resources to a minimum, we use various methods in order to receive an optimal result:

   I. *LoRA and its impact*

   While LLMs have tremendous potential, they require huge computing resources to train, meaning that only a small group of technology giants and research groups are able to build their own LLMs. How can the rest of the world create specific LLM tools to suit their needs? That's where LLM tuning comes in.

   ***LoRA (Low-Rank Adaptation)*** is a highly efficient method of LLM fine tuning, which is putting LLM development into the hands of smaller organizations and even individual developers. This makes it possible to run a specialized LLM model on a single machine, opening major opportunities for LLM development in the broader data science community.

   The [8] paper on Low-Rank Adaptation (LoRA) we find a novel approach for fine-tuning large pre-trained language models while significantly reducing the number of trainable parameters. The key idea is to freeze the original model weights and introduce low-rank matrices into the architecture, allowing for efficient adaptation to downstream tasks. It can reduce the number of trainable parameters by up to 10,000 times compared to traditional fine-tuning methods, making it feasible to run multiple experiments in parallel on limited hardware. The method also decreases GPU memory usage by approximately 3 times, which is crucial for researchers with limited computational resources.

   It modifies the fine-tuning process by freezing the original model weights and applying changes to a separate set of weights, which are then added to the original parameters. LoRA transforms the model parameters into a lower-rank dimension, reducing the number of parameters that need training, thus speeding up the process and lowering costs.

   *Advantages of Using LoRA for Fine-Tuning:*

   - ***Efficiency in Training and Adaptation:*** LoRA enhances the training and adaptation efficiency of large language models like OpenAI's GPT-3 and Meta's LLaMA. Traditional fine-tuning methods require updating all model parameters, which is computationally intensive. LoRA, instead, introduces low-rank matrices that only modify a subset of the original model's weights.
   - ***Reduced Computational Resources Requirement:*** LoRA reduces the computational resources required for fine-tuning large language models. By using low-rank matrices to update specific parameters, the approach drastically cuts down the number of parameters that need to be trained.
   - ***Preservation of Pre-trained Model Weights***: LoRA preserves the integrity of pre-trained model weights, which is a significant advantage. In traditional fine-tuning, all weights of the model are subject to change, which can lead to a loss of the general knowledge the model originally possessed. LoRA's approach of selectively updating weights through low-rank matrices ensures that the core structure and knowledge embedded in the pre-trained model are largely maintained.

   We will be using Low Rank Adaptation in scaling the Large Language Model according to limited usage of memory. This method should lower the load on the GPU usage. In conclusion, Low Rank Adaption (LoRA) presents a significant advancement in the field of natural language processing, offering a practical solution for adapting large models while minimizing resource consumption. Its empirical validation and potential for broader applications make it a valuable addition to the literature on efficient model training and adaptation.

   II. *RAG Integration*

   ***RAG Information Retrieval***: Retrieval-Augmented Generation (RAG) is a Generative AI (GenAI) framework that enhances Large Language Models (LLMs) by enabling them to access and use up-to-date and trustworthy information from internal knowledge bases and enterprise systems – without the need for retraining. It searches relevant information and queries relevant enterprise data based on the user's prompt, integrates this information into an enhanced prompt, and then invokes the LLM to deliver a more accurate and relevant response. [9] It is a novel approach that integrates retrieval mechanisms with generative models to enhance performance in knowledge-intensive tasks.

   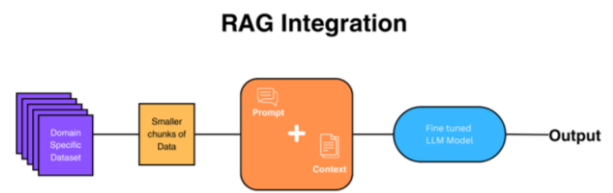

   *Fig. RAG Integration*

   It combines a pre-trained sequence-to-sequence model (parametric memory) with an external knowledge source (non-parametric memory), such as Wikipedia, to provide factual grounding. Advantages of RAG are as follows:

   1. ***Factual Accuracy***: By grounding responses in external knowledge, RAG models produce more

reliable outputs compared to traditional generative models.
2. *Flexibility*: The architecture allows for easy updates to the knowledge base, making it adaptable to new information without retraining the entire model.
3. *Diverse Applications*: RAG can be employed in various domains, including healthcare, customer service, and education, enhancing the effectiveness of information retrieval and generation tasks.

*RAG* is most useful when you need your LLM to base its responses on large amounts of updated and specifically contextual data. For example: Using RAG-based learning can dramatically enhance the educational experience by offering students access to answers and context-specific explanations based on topic-specific study materials. Also, RAG augments language translations because it enables LLMs to grasp text context and integrate terminology and domain knowledge from internal data sources.

### III. RAFT: Combining RAG with fine-tuning

Retrieval aware fine-tuning (RAFT) combines RAG and fine-tuning and provides a training recipe that improves the model's ability to answer questions in an 'open-book' domain setting. In this post, we'll explore what RAFT is and how it helps take the training of language models to the next level.

The idea is to train the model to get better at giving the right answers using what it already knows from earlier training or what it learns while being fine-tuned. Once trained, this model can also be used with additional documents to help it find answers (RAG). Here's a simple way to see how it works:

- *Training:* The model learns to go from a question to an answer (Q → A).
- *Testing without extra info* (0-shot Inference): It uses what it learned to answer new questions (Q → A).
- *Testing with RAG* (RAG Inference): It gets extra documents to help answer questions (Q+D → A).

Retrieval aware fine-tuning (RAFT) offers a new way to set up training data for models, especially for domain-specific open-book scenarios similar to in-domain RAG. In RAFT, we create training data that includes a question (Q), some documents (Dk), and a corresponding chain-of-thought (CoT) answer (A*) that's based on information from one of the documents (D*). We distinguish between two types of documents: the 'oracle' documents (D*) that have the information needed for the answer, and 'distractor' documents (Di) that don't help with the answer.

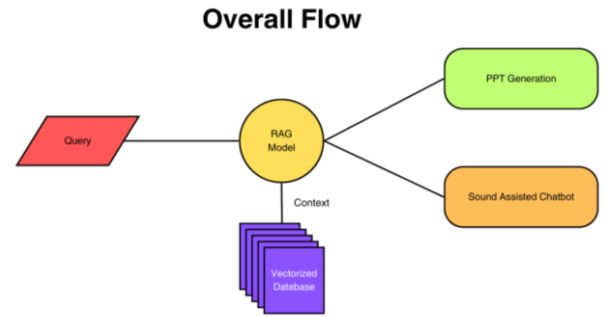

*Fig. Combination of RAG and Fine-tuning*

As we can see above, combining both and creating a RAFT with help us achieve accurate data entry as well as precise answers from the fine-tuned Large Language Models (LLMs).

### IV. Best Choice Selection

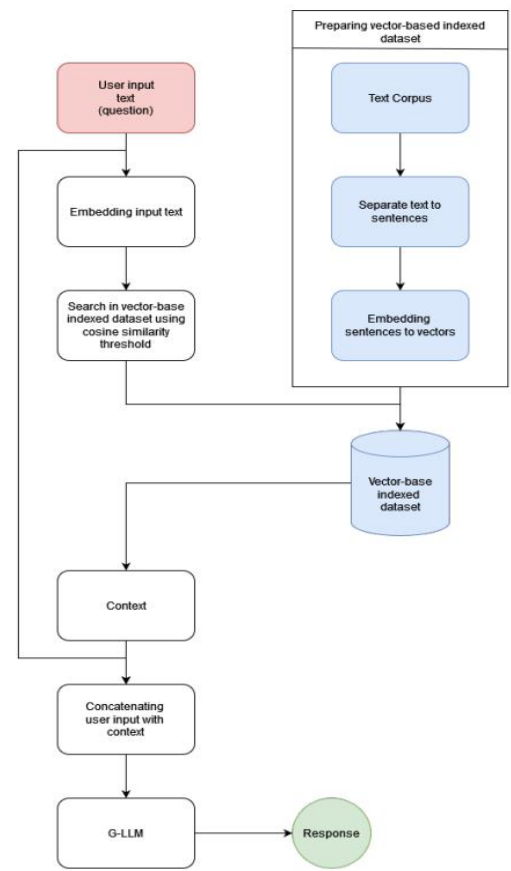

*Fig. Flow diagram of the RAG model (best approach) [10]*

[10] The research focused on Generative Large Language Models (G-LLM) models that are scientifically documented, pre-trained, and well-implemented. The thesis is organized into five sections: 1. Databases: Overview of the datasets used for parameter optimization. 2. Methodology: Description of the methodological approach for evaluating RAG and FN. 3. Measurement Results: Presentation of performance metrics for different solutions. 4.

Conclusions: Summary of findings and implications for future research. Key Findings to note are as follows:

1. ***Performance Metrics***: RAG demonstrated superior performance with 16% better ROUGE, 15% better BLEU, and 53% better cosine similarity scores compared to FN.
2. ***Risk of Hallucination***: Fine-Tuning (FN) models exhibited a higher risk of generating hallucinations, while RAG provided more reliable outputs.
3. ***Ensemble Approaches***: The potential for combining RAG and FN was explored, but it was noted that this could lead to decreased performance.

The research highlights the need for clear guidelines on building knowledge-based systems using Generative Large Language Models (G-LLMs), addressing a gap in current best practices. The findings suggest further exploration of ensemble methods and the development of more robust frameworks for integrating RAG and Fine-Tuning (FN).

This detailed conclusion provides a comprehensive overview of the research in [10], emphasizing the methodologies, findings, and implications for future studies in the field of Generative Large Language Models (G-LLM)-based knowledge systems.

### C. Performance

[10] From this study, we developed a Retrieval Augmented Generation (RAG) system designed to generate detailed responses to user queries by combining a sentence embedding model (*sentence-transformers/all-mpnet-base-v2*), a text generation model (*google/flan-t5-base*), and a FAISS-based index for fast retrieval of relevant documents. The dataset used for the test will consist of question and answers similar to training dataset. Various metrics can be calculated in order to provide efficiency such as:

*ROUGE Scores*: ROUGE (Recall-Oriented Understudy for Gisting Evaluation) metrics evaluate the overlap between the generated responses and the reference summaries.

*BLEU Scores*: BLEU (Bilingual Evaluation Understudy) metrics measure the quality of machine-generated text by comparing it to reference summaries. We evaluated BLEU-1, BLEU-2, BLEU-3, and BLEU-4 scores. This BLEU score shows a meaningful alignment between the generated and reference responses.

*Cosine Similarity*: We calculated the cosine similarity between the generated and reference summaries to measure semantic similarity, achieving a high similarity score of **0.783**. This score suggests that the generated responses maintain semantic closeness to the reference summaries.

*Hallucination Rate*: The hallucination rate, which measures the system's tendency to produce factually incorrect or irrelevant information, was calculated at ***0.99***, indicating that nearly 99% of the generated responses were relevant to the retrieved documents. This low hallucination rate demonstrates the system's reliability in generating factually consistent content.

| *Metrics* | *Result* |
|---|---|
| *ROUGE-1* | 0.442 |
| *ROUGE-2* | 0.300 |
| *ROUGE-L* | 0.381 |
| *Average BLEU* | 0.171 |
| *Cosine Similarity* | 0.783 |

*Table 1 Performance Metrics for RAG*

Overall, the RAG system presented in this [10] study represents a significant advancement in the field of retrieval-augmented text generation, offering a robust and efficient approach for producing informative and reliable responses to user queries.

### D. Saving the model

Finally, the model prepared from RAG can be saved and be ready for use. We will be using this model in order to create two types of models:

1. ***Text Generation***: This model will be made to generate context specific and accurate data in order to create a Power Point Presentation like template to demonstrate learning.
2. ***Text Summarization***: This model will be responsible for the production of short summary in order to provide it to the virtual professor for explanation through voice.

This will complete the part of overall fine-tuning the models for the usage of presentation demonstration and speaking task.

### E. Presentation Generation

Now that we have our models ready to use, we can integrate them logically in order to perform their tasks. The presentation is not an actual power point presentation but just a text generation and combination with visual demonstration. We will be using the text generation model here to generate a flow of data to insert into each slide.

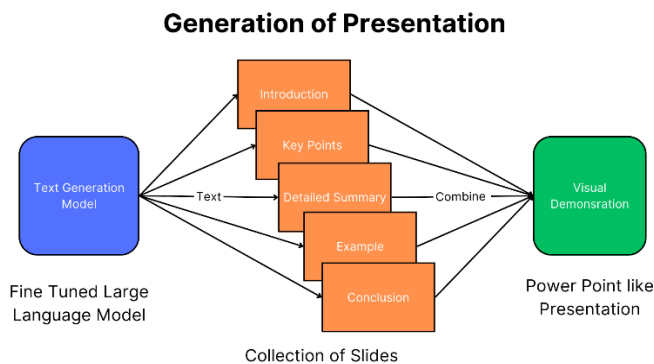

*Fig. Text Generation Model*

From the above diagram, we can understand that all the text will be coming in batches in order to get all the required information on a topic. This collection of data then will be combined in order to display as a power point presentation. The logic here is only to generate an accurate and precise template that can be used to study adequate data with explanation.

Finally, additional features and functionalities will be integrated for user convenience and tracking progress. Each topic will be designed to generate the presentation after the completion of previous topics. A user will be able to download the Presentation if needed and the generated text in presentation will be saved in the database under the user in order to avoid new data mixing or slide changes in text.

### F. Virtual Professor Integration

This part is designed to create a virtual interaction between the machine and the user in order to feel more realistic and seriousness in the environment. The professor will be a 3D model made to speak and clear user doubts. The 3D model will use the second fine tuned model that is Text Summarization model. Before we get into how the 3D model looks and works, lets understand how will the professor get the information to speak.

Having a virtual assistance that explains the modules of a specific interest makes the lectures more interactive and fun to pursue. The concept of virtual professor is designed to give a user a virtual guide that will help solve his context specific queries 24/7 without any distractions or emotional attachments. A display of physical persona will also help the user mentally create an image of actual professor to follow proper self-discipline and management in studying.

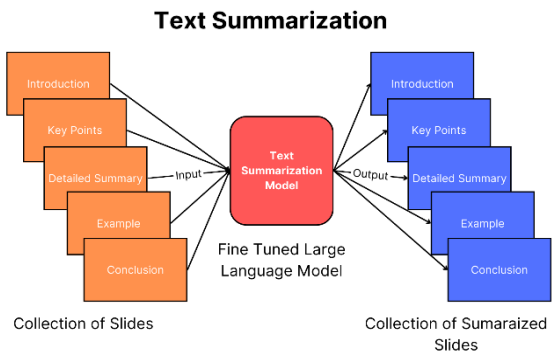

*Fig. Text Summarization Model*

In the last case, when we were generating presentation, we first created a collection of slides with different tasks allotted to each. Here, we use those slides to summarize the long paragraphs and convert it into more listenable format. This will make the output as summarized collection of slides which will become ready to be processed for the 3D model.

Now that we got the data, we now use Text-to-Speech Conversion in order to make the 3D model speak. This can be achieved by creating another model but for now, we are using Eleven Labs Application Programming Interface (API) in order to transfer the slide data as an input and get voice generated as an output.

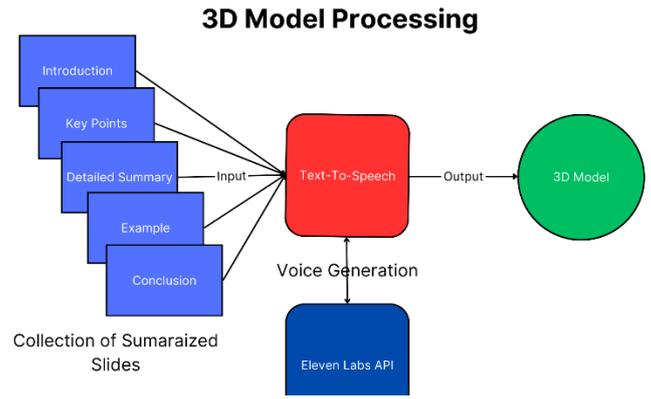

*Fig. Virtual Professor flow of work*

In order to keep consistency, we will be partially feeding the data to call the API as this will help the model speak generated text when available and meanwhile process next batch simultaneously.

The 3D Model will also be capable of solving users doubts. The doubts will be solved through voice explanation. We utilize the text summarization model here in order to implement this logic of solving user queries. This will be implemented using two ways:

1. *Chatbot integration*: A text input will be provided in the screen and it will be used to solve any context specific doubt by explanation via speech.
2. *Voice Assistance*: Using a keyword, the voice input can be trigger and by applying the same logic as the chatbot, we can send the voice input as question to the fine-tuned large language model and receive a summarized text through text-to-speech method.

### G. Roadmap Interaction

The roadmap will be designed by referencing the important topics that needs to be covered. Using a tree structure, the roadmap will be generated and each node will have a topic attached to it. The topic may have sub topics depending on the depth of it. The user has to start from the starting node and make their way towards the last node in the tree. Each node is built to work in such a way that after completion of current node, next node shall be unlocked. This way is being created in order to provide the user with more interactive and fun way of learning.

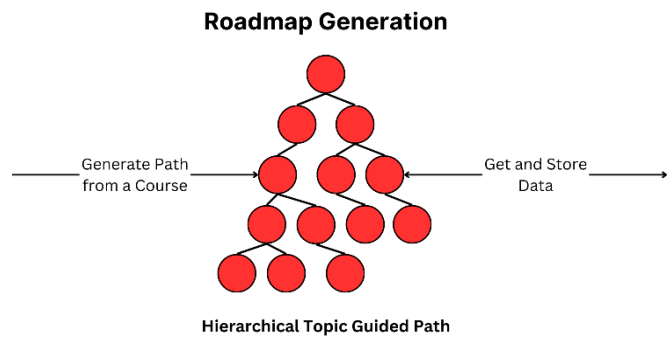

*Fig. About Roadmap Working*

After each node, user progress will be updated in the database and they can start again from where they had left before. Additionally, the text generated for presentation will also be stored under the assigned node. This way the user can have a consistency of learning previous nodes without worrying about change in data each time the node is clicked.

*H. Examination and Notes*

The architecture is focused on providing a user with authenticity and quality of resources. So, after each topic completion, the user will be tested on the basis of the knowledge gained on current topic. To check a user's level of understanding, they will be assigned with a quiz after each topic completion to get insights on the performance of learning. Based on the performance, they will be considered eligible to learn next topic. This quiz will consist of MCQs that are covered on the topic during the lecture.

Each course will have a personalised note generated using the fine-tuned Large Language Model. These notes are designed to help the user throughout the course by giving proper and adequate amount of question and answers. These can be accessed throughout the period of learning the course and also be downloaded.

Lastly, A test will be conducted at the end of each course. This test will be random and based on overall understanding of the course. The format of answering the questions will be based on long answers. The logic implemented here will be checking of answers and comparing the matching using Natural Language Processing (NLP). This can be done using Contextual Text Similarity measurement.

***Contextual text similarity*** refers to the measurement of how closely related or similar two pieces of text are in meaning, despite variations in structure, wording, or length. Traditional methods often relied on static techniques like cosine similarity or Jaccard index, which treat texts as bags of words, disregarding the context in which words appear. However, deep learning has unlocked the potential to capture contextual nuances, revolutionizing the approach to text similarity. There are basically two different ways to find similarity between texts:

1. *Lexical Similarity*: provides a measure of the similarity of two texts based on the closeness of the word sets of same or different languages such as Cosine Similarity, Jaccard Similarity, Sørensen–Dice coefficient, Levenshtein Distance.

2. *Semantic Similarity*: Semantic Similarity creates a quantitative measure of the likeness of meaning between two words or phrases such as Word/Sentence embeddings, Contextual language models, sentence transformers.

BERT with Cosine similarity: BERT is a transformers model pretrained on a large corpus of English data in a self-supervised fashion. This means it was pretrained on the raw texts only, with no humans labelling them in any way (which is why it can use lots of publicly available data) with an automatic process to generate inputs and labels from those texts.

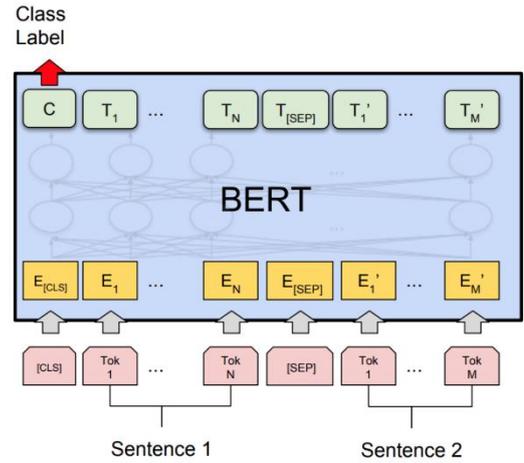

*Fig. BERT Architecture*
Source: [Pytorch Documentation](Pytorch Documentation)

*BERT Base Uncased* is a Pretrained model on English language using a masked language modeling (MLM) objective. It was introduced in this [12]BERT paper (2018). This model is uncased which means it does not make a difference between upper or lowercase, for example, "english" = "English". To tackle this, we will be processing the user input in lowercase to resolve this conflict.

This model specializes in comparing contextual information and detecting a threshold. Based on comparison, due to larger context awareness, this model helps to achieve greater and accurate results. All other methods lack context awareness to some extend and hence are avoided. Using this model, a user can be assessed more precisely and help generating specific feedback with detailed description on output.

## VII. CONCLUSION

This research paper provides a comprehensive exploration of advancements in Generative Large Language Models (G-LLMs), particularly focusing on their application in enhancing reasoning abilities and user interaction within educational frameworks. It emphasizes the importance of context awareness, as demonstrated by the BERT model, which processes input in a case-agnostic manner to improve accuracy in understanding user queries. The findings suggest that traditional methods often lack the contextual depth necessary for precise output generation, thereby underscoring the necessity for innovative approaches like Retrieval Augmented Generation (RAG). This method not only enhances the quality of generated responses but also mitigates the risk of hallucinations, a common challenge faced by fine-tuned models.

Moreover, the paper delves into the integration of advanced methodologies such as Learning by Teaching (LbT) and regularization techniques, which collectively contribute to the enhancement of model performance. The literature review articulates the theoretical foundations and empirical results, while also acknowledging the limitations and ethical considerations inherent in deploying these technologies. The exploration of LoRA (Low-Rank Adaptation) as a fine-tuning method highlights its potential to democratize access to LLM development, enabling smaller organizations and individual developers to create specialized models tailored to specific needs.

In addition, the paper outlines a structured approach to user engagement through the development of a virtual professor, which aims to facilitate a more interactive and personalized learning experience. This 3D model, capable of natural speech and doubt resolution, represents a significant step towards integrating technology in education, allowing learners to navigate complex knowledge landscapes effectively. The proposed system not only tracks user progress but also ensures the authenticity and quality of resources, thereby fostering a self-regulated learning environment. Overall, the findings contribute valuable insights to the field, paving the way for future advancements in model performance and societal impact, ultimately transforming the educational landscape into a more efficient and learner-centric domain.

At the end, all the above findings were created, tested and combined in order to form a system that will enhance a student's ability of learning broader concepts with precise details. It is always easier to find appropriate knowledge but creating new requires self-awareness. With this system, now the student will become capable of self-learning and searching their path of interests with ease.